\documentclass[%
 reprint,
 amsmath,amssymb,
 aps,
 prl,
]{revtex4-2}

\usepackage{graphicx}
\usepackage{dcolumn}
\usepackage{bm}
\usepackage{color}
\usepackage{ulem}
\usepackage{hyperref}

\begin{document}

\preprint{APS/123-QED}

\title{Machine learning identifies nullclines in oscillatory dynamical systems}

\author{Bartosz Prokop}
\author{Jimmy Billen}
\author{Nikita Frolov}
\thanks{Authors contributed equally.}
\author{Lendert Gelens}
\thanks{Authors contributed equally.}

\affiliation{
 Laboratory of Dynamics in Biological Systems,\\ Department of Cellular and Molecular Medicine, \\KU Leuven, Leuven, Belgium
}
%



\date{\today}

\begin{abstract}
We introduce CLINE (Computational Learning and Identification of Nullclines), a neural network-based method that uncovers the hidden structure of nullclines from oscillatory time series data. Unlike traditional approaches aiming at direct prediction of system dynamics, CLINE identifies static geometric features of the phase space that encode the (non)linear relationships between state variables. It overcomes challenges such as multiple time scales and strong nonlinearities while producing interpretable results convertible into symbolic differential equations. We validate CLINE on various oscillatory systems, showcasing its effectiveness.

\end{abstract}


\maketitle

\paragraph*{Introduction~--~} Dynamical systems shape the world around us, from planetary motions and weather patterns to the periodic division of cells during the cell cycle. Understanding and describing such systems allows us to explain key aspects of our world. For centuries, scientists have followed a fundamental approach: observing, collecting, and analyzing data to uncover underlying principles, which are then formulated into models based on established scientific methods and intuition \cite{Levine2024, Prokop2024}.

However, recent technological advancements in data acquisition led to the collection of vast, highly complex, and often inextricable datasets, making model derivation increasingly challenging or even impossible \cite{Miller1956, Leonelli2019}.
At the same time, modern computational advancements have driven the development of methods capable of extracting meaningful information about the underlying dynamical systems directly from measured data, requiring minimal prior knowledge.

Data-driven methods, more commonly known as machine learning, have fundamentally changed the study of dynamical systems.
These methods can be categorized based on their mode of operation and the type of output they produce:

(i) \textit{Black-box methods} are powerful tools capable of precisely replicating the system's dynamical behavior by training on large, structured datasets.
By introducing perturbations, these methods can effectively describe and predict the system behavior.
However, a key drawback is their lack of interpretability: while they provide high-level descriptions, they do not reveal the underlying interactions within the system. 
Examples of black-box methods include recurrent neural networks \cite{Penkovsky2019, Maass2002}, reservoir computing \cite{Pathak2017, Pathak2018, Weng2019, Haluszczynski2019} or neural ordinary differential equations \cite{chen2018neural, jia2019neural, dupont2019augmented, rubanova2019latent, reisser2021nodes}. 

(ii) In contrast, \textit{white-box methods} aim to represent dynamical behavior explicitly through symbolic differential equations.
 This approach allows for direct interpretation and analysis of the underlying interactions governing the system. 
However, white-box methods require high-quality data, extensive prior knowledge, and a well-founded hypothesis of possible interactions to construct a valid model \cite{Prokop2024}. 
Examples include regression-based methods such as Nonlinear Autoregressive Moving Average Model with Exogenous Inputs (NARMAX) \cite{billings2013nonlinear} and Sparse Identification of Nonlinear Dynamics (SINDy) \cite{Brunton2016}, as well as evolutionary algorithms like Symbolic Regression (SR) \cite{Schmidt2009}.

To enhance interpretability while using the strengths of deep learning, several \textit{grey-box} methods have been proposed. These approaches either integrate domain knowledge into neural network architectures—such as Physics-Informed Neural Networks (PINNs) \cite{Raissi2019, Karniadakis2021} and Biology-Informed Neural Networks (BINNs) \cite{Yazdani2020,Lagergren2020} —or translate neural network structures into mathematical descriptions of dynamical systems. The latter can take the form of complete representations (e.g., Symbolic Deep Learning \cite{Cranmer2019}) or partial formulations (e.g., Universal Differential Equations \cite{Rackauckas2020, Philipps2024})

\begin{figure}
    \centering
    \includegraphics[width=1\linewidth]{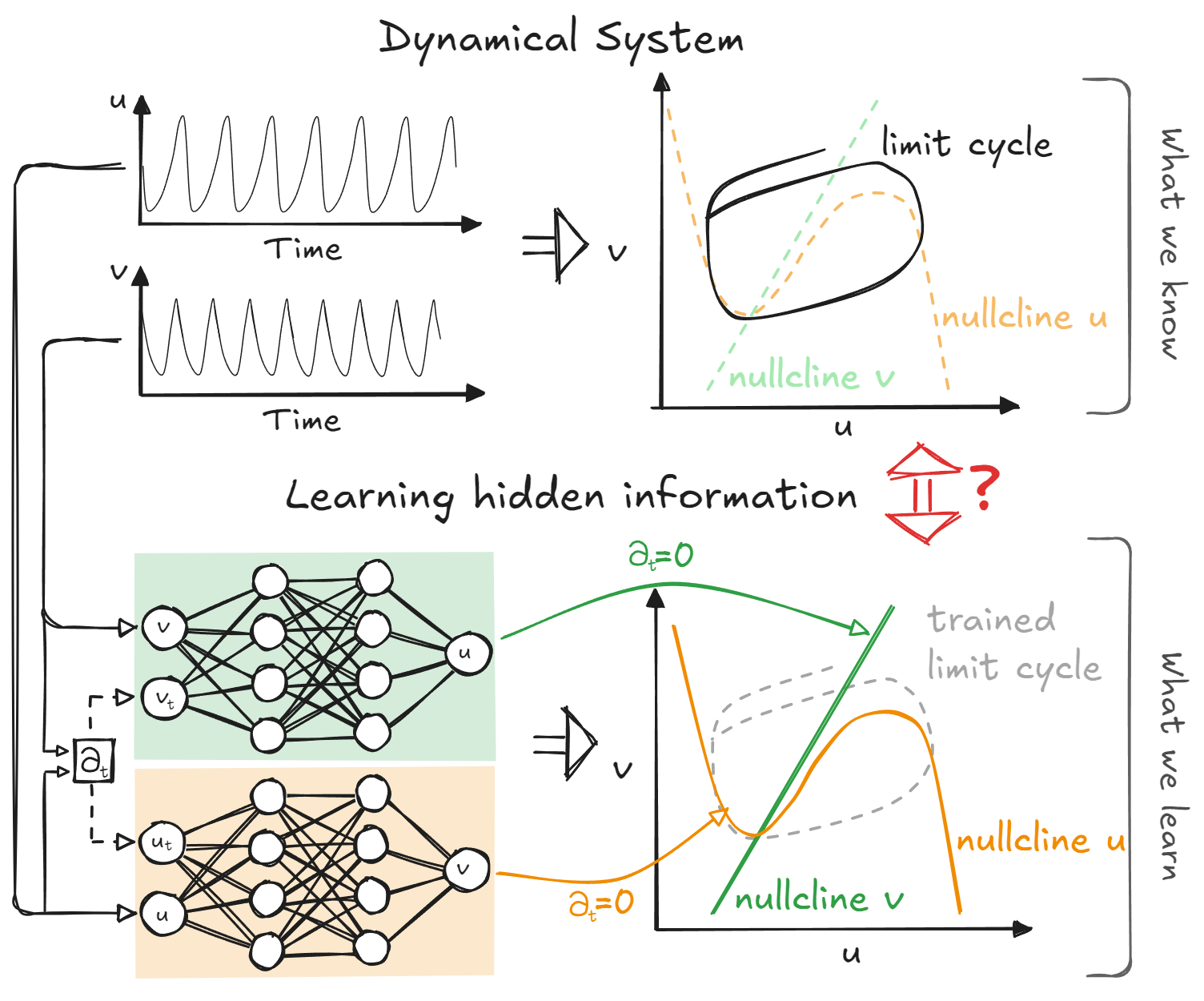}
    \caption{\textbf{CLINE uses neural networks to extract hidden nullcline structures from time series data.} Feed-forward neural networks are trained on known time series data and the derivative of a state variable of interest (e.g., $u, u_t$ or $v, v_t$) to predict the other variable ($v$ or $u$). Setting the derivative input to zero allows the network to learn the underlying nullcline structure, $f(u,v)$ or $g(u,v)$. This information can then be transformed into symbolic equations using methods such as SINDy\cite{Brunton2016}.}
    \label{fig:overview_prediction}
\end{figure}

In this work, we introduce CLINE (\textbf{C}omputational \textbf{L}earning and \textbf{I}dentification of \textbf{N}ullclin\textbf{E}s), a new method that leverages deep learning to extract interpretable models by uncovering information hidden in temporal data. Precisely, instead of focusing on forecasting dynamical systems, as done in previous studies, our approach aims to predict static geometric features in phase space, with a particular emphasis on oscillatory systems (see Fig.~\ref{fig:overview_prediction}).
CLINE identifies key attributes such as the shape of nullclines and the locations of fixed points, which serve two main purposes: (1) providing crucial constraints for symbolic model identification and (2) revealing nonlinear relationships between state variables. 

\paragraph*{The methodology ~--~}
\begin{figure*}
    \centering
    \includegraphics[width=1\linewidth]{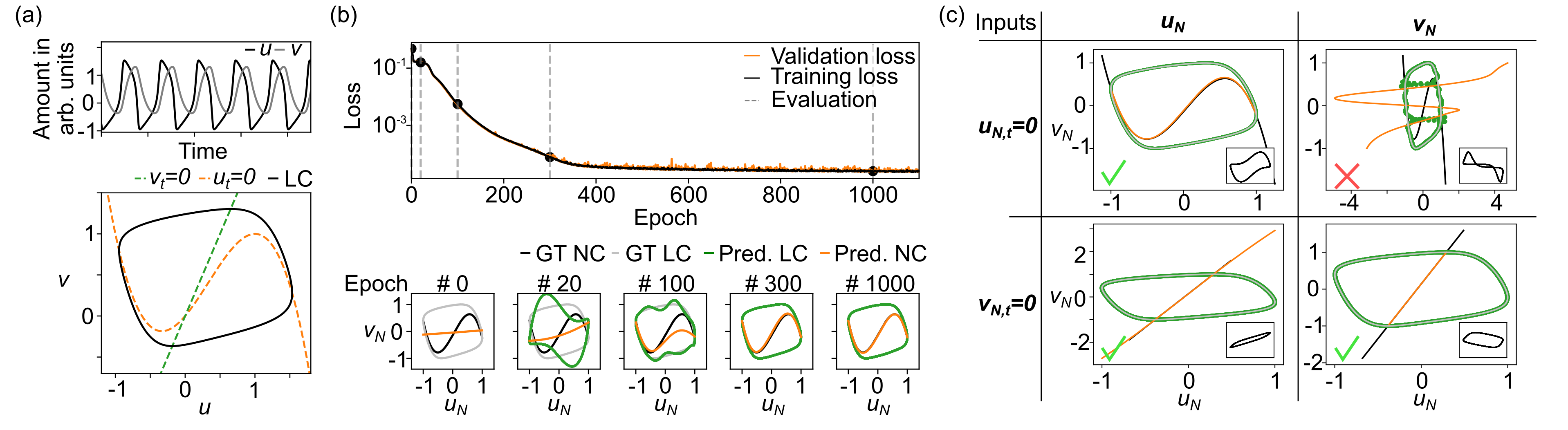}
    \caption{\textbf{Nullcline structures are accurately reconstructed when using appropriate input variables.}
(a) A representative time series of the FHN model alongside its phase portrait. The model consists of a linear nullcline ($v_t = g(u,v)$) and an S-shaped cubic nullcline ($u_t = f(u,v)$).  
(b) After normalizing the state variables, the training process enables the model to converge to the ground-truth (GT) limit cycle (LC), as demonstrated by the reconstructed S-shaped nullcline (NC).  
(c) Prediction success depends on the choice of input variables for training the deep learning model. If unsuitable inputs are used (e.g., $v$ and $u_t$ to infer $f(u,v)$), CLINE fails to correctly identify the GT limit cycle and nullcline due to poorly separated input variable limit cycles, as shown in the insets. Conversely, with appropriate input variables, nullcline identification is successful.
}
    \label{fig:FHN}
\end{figure*}
To identify the nullcline structure of an oscillatory system using CLINE, we only require measured time series of the relevant state variables. For instance, in Fig.~\ref{fig:overview_prediction}, these variables are $u$ and $v$, representing the full set of state variables. 

The dynamical behavior of such a system is fully described by a set of ordinary differential equations (ODEs):
\begin{align}
\begin{split}
    u_t = f(u,v),\\
    v_t = g(u,v).
    \label{eq:ODEs}
\end{split}
\end{align}
In general, the functional form of these equations is not defined, and $f(u,v)$ and $g(u,v)$ can, in principle, be chosen freely. However, in realistic scenarios, they follow certain rules \cite{Novak2008}: 

An oscillatory system requires a negative feedback to reset the system, as well as sufficient time delay and nonlinearity in interactions to prevent settling into a stable steady state. 
These properties can give rise to a unique and identifiable phase space structure that determines both the shape of trajectories and the attractor. 
A key determinant of the system’s phase space dynamics is the structure of the nullclines or isolines, defined as:
\begin{align}
\begin{split}
    u_t = f(u,v)=0,\\
    v_t = g(u,v)=0.
\label{eq:null_def}
\end{split}
\end{align}
This nullcline structure is not directly apparent from the attractor shape or the time series alone. However, using CLINE, we can reformulate the system of ODEs in Eq.~\eqref{eq:ODEs} to enable nullcline identification.

To this end, we express the relationship between measured variables $u$ and $v$ in terms of the inverse functions $f_{u,v}^{-1}$ and $g_{u,v}^{-1}$:
\begin{align}
    \begin{split}
        u_t &= f(u,v) \rightarrow u = f_{u}^{-1}(v,u_t)\text{ or } v = f_{v}^{-1}(u,u_t)\\
        v_t &= g(u,v) \rightarrow u = g_{u}^{-1}(v,v_t)\text{ or } v = g_{v}^{-1}(u,v_t).
    \end{split}
\label{eq:null_inverse}
\end{align}
Using this formulation, we use a deep learning algorithm to approximate $f_{u,v}^{-1}$ and $g_{u,v}^{-1}$, training the model to predict one variable based on the other one and one of the derivatives, e.g., on input data such as $u$ and $u_t$ to learn $v$
(see Fig.~\ref{fig:overview_prediction}). 

To retrieve the nullcline after training the deep learning model, we set the derivative inputs, such as $u_t$ for $f_{u,v}^{-1}$, to zero following Eqs.~(\ref{eq:null_def})-(\ref{eq:null_inverse}). 
Once trained, varying one of the input state variables (e.g., $u$) allows us to learn the corresponding $v$ variable at which $u_t=0$, thereby revealing the nullcline structure $f(u,v)$.
The required deep learning model is relatively simple as we use a feed-forward neural network. We systematically explored the influence of various network parameters, including the choice of activation functions, the number of nodes, and the network depth. Details of this analysis can be found in Supplementary Material B. 

\paragraph*{Proof of concept using the FitzHugh-Nagumo model ~--~} To demonstrate the capabilities of CLINE, we apply it to a set of generic oscillator models with different nonlinearities, following a similar approach as in Ref.~\cite{Prokop2024b}.  

We begin with the widely used FitzHugh-Nagumo (FHN) model, originally developed to describe neuronal excitability and now broadly applied across various fields, including cardiology, cell cycle modeling, electronic circuit design, and all-optical spiking neurons \cite{Cebrian-Lacasa2024}: 
\begin{align}
    \begin{split}
        \label{eq:fhn}
        u_t &= f(u,v) = -u^3 + cu^2 + du - v,\\ 
        v_t &= \varepsilon g(u,v) = \varepsilon(u - bv + a), 
    \end{split}
\end{align}
along with its corresponding nullclines (parameter values are provided in the Supplementary Materials):  
\begin{align}
    \begin{split}
        \label{eq:fhn_nullcline}
        0 &= f(u,v) = -u^3 + cu^2 + du - v,\\ 
        0 &= g(u,v) = (u - bv + a). 
    \end{split}
\end{align}
This model consists of a linear nullcline $g(u,v)$ and an S-shaped nullcline $f(u,v)$ (see Fig.~\ref{fig:FHN}(a)).

To facilitate the identification of these nullclines and improve training performance, we use min-max normalized state variables $u_N$ and $v_N$\footnote{For the FHN model and bicubic model, we normalize such that $u_N\in[-1,1]$ and $v_N\in[-1,1]$ to preserve the system's point symmetry. For other models where all values are non-negative, we normalize to $u_N\in[0,1]$ and $v_N\in[0,1]$. In general, normalization does not affect CLINE’s predictions but improves convergence.}. To determine the nullcline $f(u,v)$, we compute the derivative $u_{N,t}$ from the normalized variable $u_N$ and train the neural network over multiple epochs (see Fig.~\ref{fig:FHN}(b)). Initially, the predicted nullcline and limit cycle overlap. As training progresses, they gradually separate, ultimately leading to an accurate identification of the nullcline structure after 1000 epochs.  
The success of CLINE's predictions depends critically on the choice of input variables. When using $f(u, v)$ as the target, CLINE can be trained with either $v$ or $u$ as input (or output), depending on the formulation. In Fig.~\ref{fig:FHN}(c), we demonstrate that an improper selection of inputs leads to failed training. Specifically, using $v$ and $u_t$ as inputs to predict $u$ does \emph{not} reproduce the correct structure of $f(u, v)$. This failure arises because the nullcline $f(u, v) = 0$ can be expressed as $v(u) = -u^3 + c u^2 + d u$, so $v(u)$ is well-defined, but the inverse $u(v)$ is ambiguous.  Indeed, when examining the phase plane of the input variables $v$ and $v_t$ (see inset in Fig.~\ref{fig:FHN}(c)), we observe a self-intersecting attractor, in contrast to other input variable combinations (see Fig.~\ref{fig:FHN}(c)). This self-intersection introduces ambiguity in the relationship between the input and output variables, preventing CLINE from correctly learning the attractor and the corresponding nullcline.  
Using such phase portraits can thus be used to identify suitable input variables. With the correct input choice, CLINE robustly predicts the accurate nullcline structures for both $f(u, v)$ (top row of Fig.~\ref{fig:FHN}(c)) and $g(u, v)$ (bottom row of Fig.~\ref{fig:FHN}(c)).

\paragraph*{Handling strong time scale separation ~--~}
An important limitation of many white-box, data-driven model discovery and identification methods is their inability to accurately reconstruct dynamical systems with strong time scale separation, even when sampling strategies are adjusted \cite{Prokop2024, Messenger2024, Champion2019}. However, many real-world oscillatory systems exhibit strong time-scale separation, making them difficult to analyze using existing data-driven approaches.  

\begin{figure}
    \centering
    \includegraphics[width=0.95\linewidth]{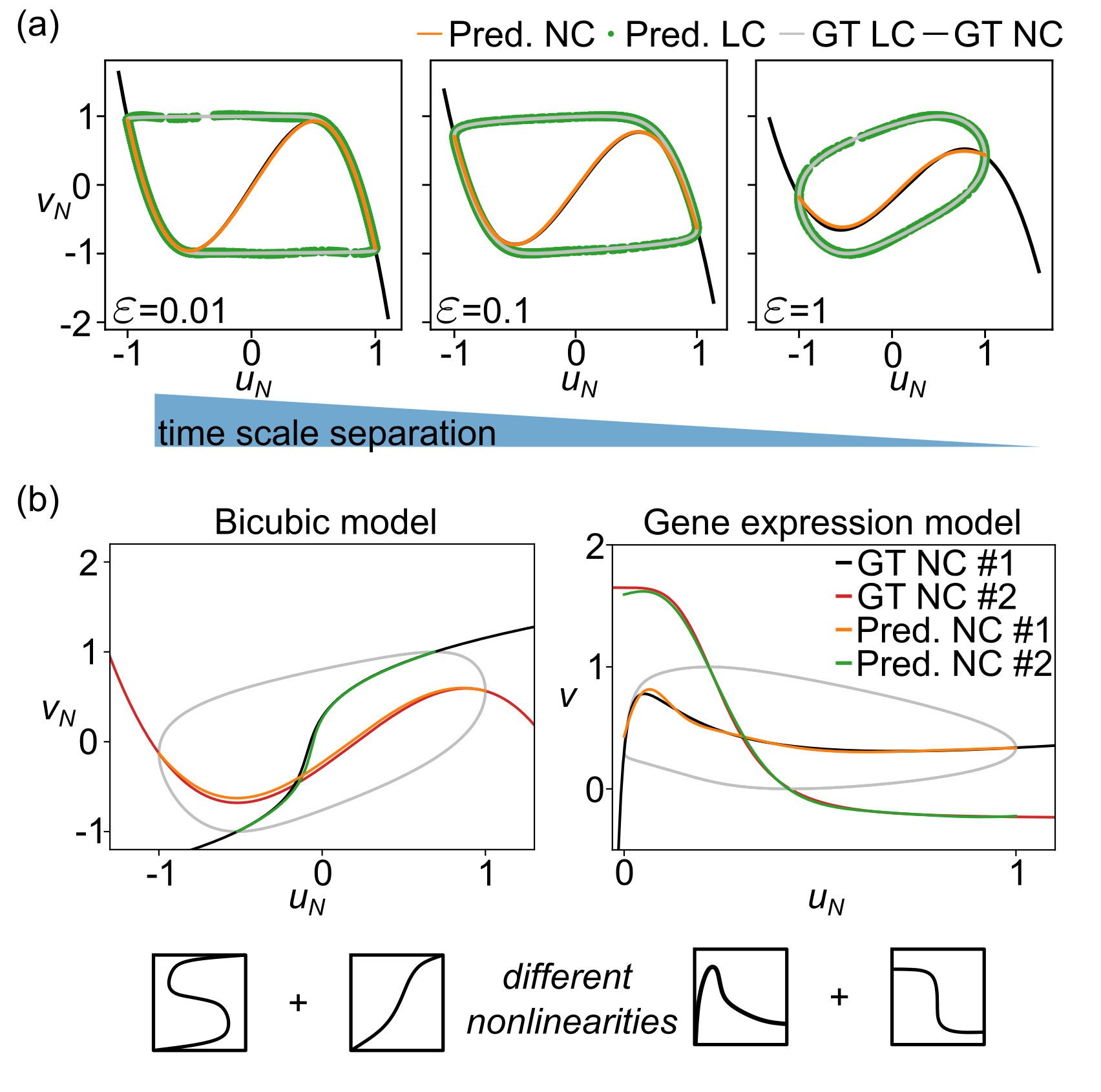}
    \caption{\textbf{CLINE robustly identifies nullclines despite time scale separation and model complexity.}
(a) In the FHN model, CLINE accurately reconstructs nullclines ($f(u,v) = 0$) across different levels of time scale separation. Stronger separation (low $\varepsilon$) improves reconstruction as the limit cycle explores more of the phase space.
(b) CLINE also generalizes to more complex models, such as the Bicubic model with two S-shaped nullclines (Eq.~\ref{eq:bicubic}) and a gene expression model with S-shaped and ultrasensitive nullclines. 
} 
    \label{fig:tss_nl_3d}
\end{figure}

As shown in Fig.~\ref{fig:tss_nl_3d}(a), we investigate CLINE’s ability to infer nullcline structures of the FHN model when varying time scale separation. The degree of time scale separation in this model is controlled by the parameter $\varepsilon$ in Eq.~(\ref{eq:fhn_nullcline}), where smaller values of $\varepsilon$ result in stronger separation and larger values lead to weaker separation. While varying $\varepsilon$ affects the shape of limit cycles and the distribution of sampled points along the fast and slow parts of the trajectory under equidistant sampling, the underlying nullclines remain unchanged (see Eq.~(\ref{eq:fhn_nullcline})).  

Many data-driven methods struggle with this sampling discrepancy across different dynamical regimes, but CLINE successfully identifies nullcline structures regardless of time scale separation (see Fig.~\ref{fig:tss_nl_3d}(a)). Interestingly, as time scale separation increases — leading to larger discrepancies in sampling — CLINE predicts the nullcline structure even more accurately. This is because a stronger separation causes the limit cycle to closely follow one of the nullclines, making reconstruction easier (illustrated for $f(u,v) = 0$ in Fig.~\ref{fig:tss_nl_3d}(a)). This robustness to time scale separation broadens the applicability of CLINE and distinguishes it from other interpretable data-driven methods with similar limitations~\cite{Prokop2024}. 

\paragraph*{Application to more complex nonlinear nullcline structures ~--~}
So far, we have demonstrated CLINE using the FHN model, which features relatively simple nullcline structures (one cubic and one linear), both expressible as polynomials. To assess its performance in more complex scenarios, we now apply CLINE to two additional models possessing different nonlinearities, as shown in Fig.~\ref{fig:tss_nl_3d}(b).

First, we consider a so-called bicubic model \cite{Prokop2024b}, which consists of two interconnected S-shaped cubic nullclines. A similar model has been shown to improve the robustness of cell cycle oscillations \cite{DeBoeck2021, Parra-Rivas2023}.
Although more complex, these nullclines can still be expressed as polynomials:
\begin{align}
\begin{split}
\label{eq:bicubic}
    u_t &= -u^3 + a u^2 +b u + c v, \\
    v_t &= d v^3 +e v^2 +f v + g u.
\end{split}
\end{align}
With appropriately chosen parameter values (see Supplementary Materials A), the bicubic model generates time series that appear visually indistinguishable from those of the FHN model, despite being governed by different nonlinearities (see Supplementary Materials C for further details).

Secondly, we apply CLINE to a gene expression model from Ref.~\cite{Novak2008}, which features an S-shaped and an ultrasensitive nullcline described by a Hill function (see Supplementary Materials A for parameters):
 
\begin{align}
    \begin{split}
        u_t &= k_1 S \frac{K^n_d}{K^n_d+v^n} - k_{du} u,\\
        v_t &= k_{sv} u - k_{dv} v - k_2 E_T \frac{v}{K_m + v + K_I v^2}.
    \end{split}
\end{align}
This model presents a significant challenge for data-driven methods that rely on predefined libraries of terms to construct symbolic equations, as both nullcline equations are highly nonlinear and rational.

However, as shown in Fig.~\ref{fig:tss_nl_3d}(b), CLINE accurately reconstructs the nullcline structures of both models, regardless of their complexity or functional form — provided the correct input variables are chosen (see Supplementary Materials D). CLINE also remains effective even when time series from different dynamical systems, such as the bicubic and FHN models, appear visually similar. Interestingly, in the gene expression model, CLINE predicts the structure of the ultrasensitive nullcline even beyond the limit cycle, suggesting that, with proper training, the method can infer nullcline structures even outside the observed attractor. However, while this result is promising, it is not generally guaranteed that CLINE will consistently extrapolate beyond the training data — an aspect that warrants further investigation in future work.

\begin{figure}
    \centering
    \includegraphics[width=1\linewidth]{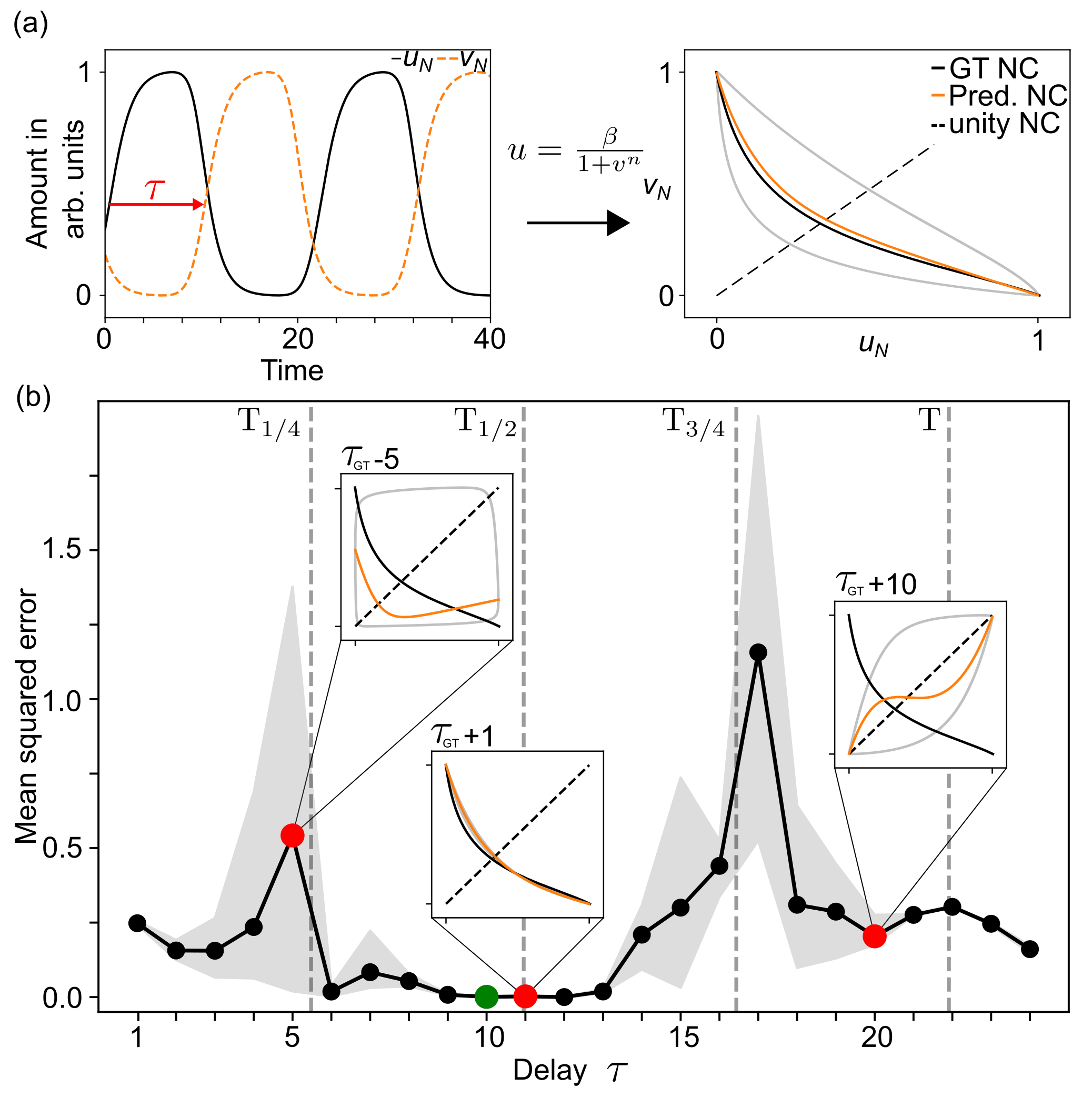}
    \caption{\textbf{CLINE accurately identifies nullclines in DDEs when the delay is close to the true value.}
(a) In the DDE from Eq. (\ref{eq:DDE}), the second variable is introduced as $v = u(t - \tau)$. This creates a nonlinear and rational nullcline well approximated by CLINE.
(b) When the chosen delay $\tau$ is close to the ground-truth delay $\tau_{\text{GT}} = 10$ (green dot), specifically within $\tau_{\text{GT}} - 1 \leq \tau \leq \tau_{\text{GT}} + 2$, the phase space and identified nullcline remain accurate with low mean squared error (MSE) and small variations. However, for larger deviations (e.g., $\tau = \tau_{\text{GT}} - 5$ or $\tau = \tau_{\text{GT}} + 10$), CLINE fails to recover the correct nullcline, leading to increased MSE (error bars over five iterations). Errors are particularly pronounced when $\tau$ corresponds to $T/4$ or $3T/4$ of the system's period $T$, as seen for $\tau = \tau_{\text{GT}} - 5$, where the limit cycle lacks distinct maxima and minima.} 
    \label{fig:dde}
\end{figure}
Finally, we turn to a special class of oscillatory systems described by delay differential equations (DDEs) explicitly incorporating a time delay $\tau$ into the model formulation. Unlike previous models, a DDE can generate oscillations with just a single variable by constructing the second necessary variable through the delay term \cite{strogatz2018nonlinear}. DDEs play a crucial role in various fields, including epidemiology \cite{Gopalsamy1992}, biology \cite{Rombouts2018,Glass2021,rombouts2023ups}, physics \cite{Ikeda1980}, and economics \cite{boucekkine2004modelling}.


\paragraph*{Prediction for delay oscillators ~--~}
We select a simple example of a DDE that models the activity of an auto-inhibitory gene (inspired by Ref.~\cite{Lewis2003}; parameter details are provided in Supplementary Material A):

\begin{align}
\begin{split}
  u_t &= \frac{\beta}{1+v^n} - u, \text{with}\\ 
   v(t) &= u(t-\tau).  
\label{eq:DDE}
\end{split}
\end{align}
Given the correct delay $\tau$, we set up CLINE similarly to its application in two-dimensional systems. However, for a DDE, only one nullcline is determined by Eq.~(\ref{eq:DDE}), while the other is always defined as $u = v$. Applied in this way, CLINE successfully identifies the nullcline structure, as shown in Fig.~\ref{fig:dde}(a).

It is important to note that the actual nullcline structure is accurately identified only when the chosen delay $\tau$ matches or is close to the ground-truth delay ($\tau_{\text{GT}}$), specifically within the range $\tau_{\text{GT}} - 1 \leq \tau \leq \tau_{\text{GT}} + 2$ (see Fig.\ref{fig:dde}(b)). Within this range, the prediction error remains small, and results across multiple realizations (here, five, shown in light gray in Fig.\ref{fig:dde}(b)) exhibit minimal variation. This suggests that even if the selected delay is slightly incorrect but still close to $\tau_{\text{GT}}$, nullcline identification can still be successful.

However, as the chosen delay deviates further from $\tau_{\text{GT}}$ (e.g., $\tau = \tau_{\text{GT}} - 5$ or $\tau = \tau_{\text{GT}} + 10$, see Fig.\ref{fig:dde}(b)), the accuracy of nullcline reconstruction becomes worse. In particular, when $\tau$ is near $1/4$ or $3/4$ of the system’s period $T$ (dashed lines in Fig.\ref{fig:dde}(b)), nullcline prediction fails and becomes highly variable. This failure arises from insufficient variable separation, introducing ambiguity in the inferred nullcline structure.

Nevertheless, even when the selected delay is too large or too small, the predicted ``incorrect'', nullcline structures still exhibit expected qualitative features: they appear nonlinear, resemble cubic equations, and have turning points at the minima and maxima of the limit cycle. Determining an appropriate delay embedding in practical scenarios is beyond the scope of this work; for further details, we refer the reader to Ref.~\cite{Tan2023}.

\paragraph*{Discussion ~--~}
In summary, this work introduces CLINE, a novel neural-network-based method that represents a fundamental change in how dynamical systems can be studied using data-driven approaches.
Rather than focusing on forecasting, we propose improving system identification by extracting phase space features — specifically, nullcline structures — from time-series data.
These structures can then be converted into symbolic equations using either data-driven techniques or traditional methods, such as SINDy or SR \cite{Brunton2016, Schmidt2009}, simplifying the explicit model identification process. We demonstrate that for dynamical systems exhibiting oscillatory behavior, CLINE accurately identifies nullcline structures while offering several advantages over existing methods.

One key advantage of CLINE is its insensitivity to time-scale separation, a common challenge in real-world dynamical systems where explicit symbolic methods often struggle \cite{Prokop2024, Messenger2024}. Additionally, CLINE effectively predicts highly nonlinear nullcline structures without relying on predefined sets of terms or interactions, making it more flexible and less biased.
We illustrate these strengths by applying CLINE to various models incorporating one or more of these challenges, focusing on systems governed by two coupled ODEs or a single DDE.

Currently, CLINE has two primary limitations.
First, it assumes access to all state variables and does not address the case of partial observations.
For instance, if only one system variable is known, embedding techniques must be applied, which may affect the accuracy of nullcline prediction.
Second, we do not explore nullcline inference for systems of dimensionality more than two.
Preliminary results suggest that in such cases, careful selection of inputs is necessary, as the directionality and clarity of input-output relationships become less straightforward compared to two-dimensional ODE systems.
Both aspects require further investigation to assess the broader applicability of CLINE.

Future work should also examine the method’s prediction horizon.
Our initial results indicate that, when adequately trained, CLINE can predict nullcline structures beyond the limit cycle's location in phase space.
However, the data quality and system conditions necessary for robust predictions outside the provided limits remain unclear.
In particular, the impact of data quality — such as sampling rates and noise levels — requires deeper analysis.
Moreover, while our study focuses on oscillatory systems, an open question remains: To what extent can invariant systems or purely transient behaviors be used for nullcline identification with CLINE?
We speculate that if sufficient transient or invariant trajectories are provided during training, CLINE should be capable of identifying nullcline structures for such systems.

Ultimately, CLINE offers a fundamentally different perspective on studying dynamical systems using data-driven or machine learning methods — shifting the focus from temporal evolution to static phase space features.
When combined with symbolic techniques, CLINE has the potential to become a powerful tool for analyzing real-world dynamical systems, generating new models, and uncovering novel insights into the underlying interactions of dynamical (oscillatory) systems.

\paragraph*{Acknowledgments ~--~} 
L.G. acknowledges funding by the KU Leuven Research Fund (grant number C14/23/130) and the Research-Foundation Flanders (FWO, grant number G074321N).

\paragraph*{Data availability ~--~}
The data and code that support the findings of this article are openly available \cite{gitlab_cline2025}.



\bibliography{bibliography.bib}

\begin{thebibliography}{43}%
\makeatletter
\providecommand \@ifxundefined [1]{%
 \@ifx{#1\undefined}
}%
\providecommand \@ifnum [1]{%
 \ifnum #1\expandafter \@firstoftwo
 \else \expandafter \@secondoftwo
 \fi
}%
\providecommand \@ifx [1]{%
 \ifx #1\expandafter \@firstoftwo
 \else \expandafter \@secondoftwo
 \fi
}%
\providecommand \natexlab [1]{#1}%
\providecommand \enquote  [1]{``#1''}%
\providecommand \bibnamefont  [1]{#1}%
\providecommand \bibfnamefont [1]{#1}%
\providecommand \citenamefont [1]{#1}%
\providecommand \href@noop [0]{\@secondoftwo}%
\providecommand \href [0]{\begingroup \@sanitize@url \@href}%
\providecommand \@href[1]{\@@startlink{#1}\@@href}%
\providecommand \@@href[1]{\endgroup#1\@@endlink}%
\providecommand \@sanitize@url [0]{\catcode `\\12\catcode `\$12\catcode `\&12\catcode `\#12\catcode `\^12\catcode `\_12\catcode `\%12\relax}%
\providecommand \@@startlink[1]{}%
\providecommand \@@endlink[0]{}%
\providecommand \url  [0]{\begingroup\@sanitize@url \@url }%
\providecommand \@url [1]{\endgroup\@href {#1}{\urlprefix }}%
\providecommand \urlprefix  [0]{URL }%
\providecommand \Eprint [0]{\href }%
\providecommand \doibase [0]{https://doi.org/}%
\providecommand \selectlanguage [0]{\@gobble}%
\providecommand \bibinfo  [0]{\@secondoftwo}%
\providecommand \bibfield  [0]{\@secondoftwo}%
\providecommand \translation [1]{[#1]}%
\providecommand \BibitemOpen [0]{}%
\providecommand \bibitemStop [0]{}%
\providecommand \bibitemNoStop [0]{.\EOS\space}%
\providecommand \EOS [0]{\spacefactor3000\relax}%
\providecommand \BibitemShut  [1]{\csname bibitem#1\endcsname}%
\let\auto@bib@innerbib\@empty
\bibitem [{\citenamefont {Levine}\ and\ \citenamefont {Tu}(2024)}]{Levine2024}%
  \BibitemOpen
  \bibfield  {author} {\bibinfo {author} {\bibfnamefont {H.}~\bibnamefont {Levine}}\ and\ \bibinfo {author} {\bibfnamefont {Y.}~\bibnamefont {Tu}},\ }\bibfield  {title} {\bibinfo {title} {{Machine learning meets physics: A two-way street}},\ }\href {https://doi.org/10.1073/PNAS.2403580121/ASSET/E6CF82CE-0779-4C75-BE4B-E261CED976B6/ASSETS/IMAGES/LARGE/PNAS.2403580121FIG03.JPG} {\bibfield  {journal} {\bibinfo  {journal} {Proceedings of the National Academy of Sciences of the United States of America}\ }\textbf {\bibinfo {volume} {121}},\ \bibinfo {pages} {e2403580121} (\bibinfo {year} {2024})}\BibitemShut {NoStop}%
\bibitem [{\citenamefont {Prokop}\ and\ \citenamefont {Gelens}(2024)}]{Prokop2024}%
  \BibitemOpen
  \bibfield  {author} {\bibinfo {author} {\bibfnamefont {B.}~\bibnamefont {Prokop}}\ and\ \bibinfo {author} {\bibfnamefont {L.}~\bibnamefont {Gelens}},\ }\bibfield  {title} {\bibinfo {title} {{From biological data to oscillator models using SINDy}},\ }\href {https://doi.org/10.1016/J.ISCI.2024.109316} {\bibfield  {journal} {\bibinfo  {journal} {iScience}\ }\textbf {\bibinfo {volume} {27}},\ \bibinfo {pages} {109316} (\bibinfo {year} {2024})}\BibitemShut {NoStop}%
\bibitem [{\citenamefont {Miller}(1956)}]{Miller1956}%
  \BibitemOpen
  \bibfield  {author} {\bibinfo {author} {\bibfnamefont {G.~A.}\ \bibnamefont {Miller}},\ }\bibfield  {title} {\bibinfo {title} {{The magical number seven, plus or minus two: Some limits on our capacity for processing information.}},\ }\href {https://doi.org/10.1037/h0043158} {\bibfield  {journal} {\bibinfo  {journal} {Psychological Review}\ }\textbf {\bibinfo {volume} {63}},\ \bibinfo {pages} {81} (\bibinfo {year} {1956})}\BibitemShut {NoStop}%
\bibitem [{\citenamefont {Leonelli}(2019)}]{Leonelli2019}%
  \BibitemOpen
  \bibfield  {author} {\bibinfo {author} {\bibfnamefont {S.}~\bibnamefont {Leonelli}},\ }\bibfield  {title} {\bibinfo {title} {{The challenges of big data biology}},\ }\bibfield  {journal} {\bibinfo  {journal} {eLife}\ }\textbf {\bibinfo {volume} {8}},\ \href {https://doi.org/10.7554/ELIFE.47381} {10.7554/ELIFE.47381} (\bibinfo {year} {2019})\BibitemShut {NoStop}%
\bibitem [{\citenamefont {Penkovsky}\ \emph {et~al.}(2019)\citenamefont {Penkovsky}, \citenamefont {Porte}, \citenamefont {Jacquot}, \citenamefont {Larger},\ and\ \citenamefont {Brunner}}]{Penkovsky2019}%
  \BibitemOpen
  \bibfield  {author} {\bibinfo {author} {\bibfnamefont {B.}~\bibnamefont {Penkovsky}}, \bibinfo {author} {\bibfnamefont {X.}~\bibnamefont {Porte}}, \bibinfo {author} {\bibfnamefont {M.}~\bibnamefont {Jacquot}}, \bibinfo {author} {\bibfnamefont {L.}~\bibnamefont {Larger}},\ and\ \bibinfo {author} {\bibfnamefont {D.}~\bibnamefont {Brunner}},\ }\bibfield  {title} {\bibinfo {title} {{Coupled nonlinear delay systems as deep convolutional neural networks}},\ }\bibfield  {journal} {\bibinfo  {journal} {Physical Review Letters}\ }\textbf {\bibinfo {volume} {123}},\ \href {https://doi.org/10.1103/PHYSREVLETT.123.054101} {10.1103/PHYSREVLETT.123.054101} (\bibinfo {year} {2019}),\ \Eprint {https://arxiv.org/abs/1902.05608} {arXiv:1902.05608} \BibitemShut {NoStop}%
\bibitem [{\citenamefont {Maass}\ \emph {et~al.}(2002)\citenamefont {Maass}, \citenamefont {Natschl{\"{a}}ger},\ and\ \citenamefont {Markram}}]{Maass2002}%
  \BibitemOpen
  \bibfield  {author} {\bibinfo {author} {\bibfnamefont {W.}~\bibnamefont {Maass}}, \bibinfo {author} {\bibfnamefont {T.}~\bibnamefont {Natschl{\"{a}}ger}},\ and\ \bibinfo {author} {\bibfnamefont {H.}~\bibnamefont {Markram}},\ }\bibfield  {title} {\bibinfo {title} {{Real-Time Computing Without Stable States: A New Framework for Neural Computation Based on Perturbations}},\ }\href {https://doi.org/10.1162/089976602760407955} {\bibfield  {journal} {\bibinfo  {journal} {Neural Computation}\ }\textbf {\bibinfo {volume} {14}},\ \bibinfo {pages} {2531} (\bibinfo {year} {2002})}\BibitemShut {NoStop}%
\bibitem [{\citenamefont {Pathak}\ \emph {et~al.}(2017)\citenamefont {Pathak}, \citenamefont {Lu}, \citenamefont {Hunt}, \citenamefont {Girvan},\ and\ \citenamefont {Ott}}]{Pathak2017}%
  \BibitemOpen
  \bibfield  {author} {\bibinfo {author} {\bibfnamefont {J.}~\bibnamefont {Pathak}}, \bibinfo {author} {\bibfnamefont {Z.}~\bibnamefont {Lu}}, \bibinfo {author} {\bibfnamefont {B.~R.}\ \bibnamefont {Hunt}}, \bibinfo {author} {\bibfnamefont {M.}~\bibnamefont {Girvan}},\ and\ \bibinfo {author} {\bibfnamefont {E.}~\bibnamefont {Ott}},\ }\bibfield  {title} {\bibinfo {title} {{Using machine learning to replicate chaotic attractors and calculate Lyapunov exponents from data}},\ }\href {https://doi.org/10.1063/1.5010300/135382} {\bibfield  {journal} {\bibinfo  {journal} {Chaos}\ }\textbf {\bibinfo {volume} {27}},\ \bibinfo {pages} {121102} (\bibinfo {year} {2017})},\ \Eprint {https://arxiv.org/abs/1710.07313} {arXiv:1710.07313} \BibitemShut {NoStop}%
\bibitem [{\citenamefont {Pathak}\ \emph {et~al.}(2018)\citenamefont {Pathak}, \citenamefont {Hunt}, \citenamefont {Girvan}, \citenamefont {Lu},\ and\ \citenamefont {Ott}}]{Pathak2018}%
  \BibitemOpen
  \bibfield  {author} {\bibinfo {author} {\bibfnamefont {J.}~\bibnamefont {Pathak}}, \bibinfo {author} {\bibfnamefont {B.}~\bibnamefont {Hunt}}, \bibinfo {author} {\bibfnamefont {M.}~\bibnamefont {Girvan}}, \bibinfo {author} {\bibfnamefont {Z.}~\bibnamefont {Lu}},\ and\ \bibinfo {author} {\bibfnamefont {E.}~\bibnamefont {Ott}},\ }\bibfield  {title} {\bibinfo {title} {{Model-Free Prediction of Large Spatiotemporally Chaotic Systems from Data: A Reservoir Computing Approach}},\ }\href {https://doi.org/10.1103/PHYSREVLETT.120.024102/APS_SUPPLEMENT.PDF} {\bibfield  {journal} {\bibinfo  {journal} {Physical Review Letters}\ }\textbf {\bibinfo {volume} {120}},\ \bibinfo {pages} {024102} (\bibinfo {year} {2018})}\BibitemShut {NoStop}%
\bibitem [{\citenamefont {Weng}\ \emph {et~al.}(2019)\citenamefont {Weng}, \citenamefont {Yang}, \citenamefont {Gu}, \citenamefont {Zhang},\ and\ \citenamefont {Small}}]{Weng2019}%
  \BibitemOpen
  \bibfield  {author} {\bibinfo {author} {\bibfnamefont {T.}~\bibnamefont {Weng}}, \bibinfo {author} {\bibfnamefont {H.}~\bibnamefont {Yang}}, \bibinfo {author} {\bibfnamefont {C.}~\bibnamefont {Gu}}, \bibinfo {author} {\bibfnamefont {J.}~\bibnamefont {Zhang}},\ and\ \bibinfo {author} {\bibfnamefont {M.}~\bibnamefont {Small}},\ }\bibfield  {title} {\bibinfo {title} {{Synchronization of chaotic systems and their machine-learning models.}},\ }\bibfield  {journal} {\bibinfo  {journal} {Physical Review E}\ }\textbf {\bibinfo {volume} {99}},\ \href {https://doi.org/10.1103/PHYSREVE.99.042203} {10.1103/PHYSREVE.99.042203} (\bibinfo {year} {2019})\BibitemShut {NoStop}%
\bibitem [{\citenamefont {Haluszczynski}\ and\ \citenamefont {R{\"{a}}th}(2019)}]{Haluszczynski2019}%
  \BibitemOpen
  \bibfield  {author} {\bibinfo {author} {\bibfnamefont {A.}~\bibnamefont {Haluszczynski}}\ and\ \bibinfo {author} {\bibfnamefont {C.}~\bibnamefont {R{\"{a}}th}},\ }\bibfield  {title} {\bibinfo {title} {{Good and bad predictions: Assessing and improving the replication of chaotic attractors by means of reservoir computing}},\ }\href {https://doi.org/10.1063/1.5118725/282708} {\bibfield  {journal} {\bibinfo  {journal} {Chaos}\ }\textbf {\bibinfo {volume} {29}},\ \bibinfo {pages} {103143} (\bibinfo {year} {2019})},\ \Eprint {https://arxiv.org/abs/1907.05639} {arXiv:1907.05639} \BibitemShut {NoStop}%
\bibitem [{\citenamefont {Chen}\ \emph {et~al.}(2018)\citenamefont {Chen}, \citenamefont {Rubanova}, \citenamefont {Bettencourt},\ and\ \citenamefont {Duvenaud}}]{chen2018neural}%
  \BibitemOpen
  \bibfield  {author} {\bibinfo {author} {\bibfnamefont {R.~T.~Q.}\ \bibnamefont {Chen}}, \bibinfo {author} {\bibfnamefont {Y.}~\bibnamefont {Rubanova}}, \bibinfo {author} {\bibfnamefont {J.}~\bibnamefont {Bettencourt}},\ and\ \bibinfo {author} {\bibfnamefont {D.}~\bibnamefont {Duvenaud}},\ }\bibfield  {title} {\bibinfo {title} {Neural ordinary differential equations},\ }in\ \href@noop {} {\emph {\bibinfo {booktitle} {Advances in neural information processing systems (NeurIPS)}}}\ (\bibinfo {year} {2018})\ \Eprint {https://arxiv.org/abs/1806.07366} {arXiv:1806.07366} \BibitemShut {NoStop}%
\bibitem [{\citenamefont {Jia}\ \emph {et~al.}(2021)\citenamefont {Jia}, \citenamefont {Chen}, \citenamefont {Duvenaud},\ and\ \citenamefont {Brubaker}}]{jia2019neural}%
  \BibitemOpen
  \bibfield  {author} {\bibinfo {author} {\bibfnamefont {J.}~\bibnamefont {Jia}}, \bibinfo {author} {\bibfnamefont {R.~T.~Q.}\ \bibnamefont {Chen}}, \bibinfo {author} {\bibfnamefont {D.}~\bibnamefont {Duvenaud}},\ and\ \bibinfo {author} {\bibfnamefont {M.~A.}\ \bibnamefont {Brubaker}},\ }\bibfield  {title} {\bibinfo {title} {Neural sdes: Stochastic differential equations as continuous-time machine learning models},\ }in\ \href@noop {} {\emph {\bibinfo {booktitle} {International Conference on Learning Representations (ICLR)}}}\ (\bibinfo {year} {2021})\ \Eprint {https://arxiv.org/abs/1905.09883} {arXiv:1905.09883} \BibitemShut {NoStop}%
\bibitem [{\citenamefont {Dupont}\ \emph {et~al.}(2019)\citenamefont {Dupont}, \citenamefont {Doucet},\ and\ \citenamefont {Teh}}]{dupont2019augmented}%
  \BibitemOpen
  \bibfield  {author} {\bibinfo {author} {\bibfnamefont {E.}~\bibnamefont {Dupont}}, \bibinfo {author} {\bibfnamefont {A.}~\bibnamefont {Doucet}},\ and\ \bibinfo {author} {\bibfnamefont {Y.~W.}\ \bibnamefont {Teh}},\ }\bibfield  {title} {\bibinfo {title} {Augmented neural odes},\ }in\ \href@noop {} {\emph {\bibinfo {booktitle} {Advances in neural information processing systems (NeurIPS)}}}\ (\bibinfo {year} {2019})\ \Eprint {https://arxiv.org/abs/1904.01681} {arXiv:1904.01681} \BibitemShut {NoStop}%
\bibitem [{\citenamefont {Rubanova}\ \emph {et~al.}(2019)\citenamefont {Rubanova}, \citenamefont {Chen},\ and\ \citenamefont {Duvenaud}}]{rubanova2019latent}%
  \BibitemOpen
  \bibfield  {author} {\bibinfo {author} {\bibfnamefont {Y.}~\bibnamefont {Rubanova}}, \bibinfo {author} {\bibfnamefont {R.~T.~Q.}\ \bibnamefont {Chen}},\ and\ \bibinfo {author} {\bibfnamefont {D.}~\bibnamefont {Duvenaud}},\ }\bibfield  {title} {\bibinfo {title} {Latent odes for irregularly-sampled time series},\ }in\ \href@noop {} {\emph {\bibinfo {booktitle} {Advances in neural information processing systems (NeurIPS)}}}\ (\bibinfo {year} {2019})\ \Eprint {https://arxiv.org/abs/1907.03907} {arXiv:1907.03907} \BibitemShut {NoStop}%
\bibitem [{\citenamefont {Reisser}\ \emph {et~al.}(2021)\citenamefont {Reisser}, \citenamefont {Maier-Hein}, \citenamefont {Kaissis}, \citenamefont {Rueckert},\ and\ \citenamefont {Feussner}}]{reisser2021nodes}%
  \BibitemOpen
  \bibfield  {author} {\bibinfo {author} {\bibfnamefont {J.}~\bibnamefont {Reisser}}, \bibinfo {author} {\bibfnamefont {L.}~\bibnamefont {Maier-Hein}}, \bibinfo {author} {\bibfnamefont {G.}~\bibnamefont {Kaissis}}, \bibinfo {author} {\bibfnamefont {D.}~\bibnamefont {Rueckert}},\ and\ \bibinfo {author} {\bibfnamefont {H.}~\bibnamefont {Feussner}},\ }\bibfield  {title} {\bibinfo {title} {Nodes as flows: A closer look at continuous normalizing flows},\ }in\ \href@noop {} {\emph {\bibinfo {booktitle} {International Conference on Learning Representations (ICLR)}}}\ (\bibinfo {year} {2021})\ \Eprint {https://arxiv.org/abs/2006.06663} {arXiv:2006.06663} \BibitemShut {NoStop}%
\bibitem [{\citenamefont {Billings}(2013)}]{billings2013nonlinear}%
  \BibitemOpen
  \bibfield  {author} {\bibinfo {author} {\bibfnamefont {S.~A.}\ \bibnamefont {Billings}},\ }\href {https://doi.org/10.1002/9781118535561} {\emph {\bibinfo {title} {{Nonlinear System Identification}}}}\ (\bibinfo  {publisher} {John Wiley \& Sons, Ltd},\ \bibinfo {address} {Chichester, UK},\ \bibinfo {year} {2013})\BibitemShut {NoStop}%
\bibitem [{\citenamefont {Brunton}\ \emph {et~al.}(2016)\citenamefont {Brunton}, \citenamefont {Proctor},\ and\ \citenamefont {Kutz}}]{Brunton2016}%
  \BibitemOpen
  \bibfield  {author} {\bibinfo {author} {\bibfnamefont {S.~L.}\ \bibnamefont {Brunton}}, \bibinfo {author} {\bibfnamefont {J.~L.}\ \bibnamefont {Proctor}},\ and\ \bibinfo {author} {\bibfnamefont {J.~N.}\ \bibnamefont {Kutz}},\ }\bibfield  {title} {\bibinfo {title} {{Discovering governing equations from data by sparse identification of nonlinear dynamical systems}},\ }\href {https://doi.org/10.1073/pnas.1517384113} {\bibfield  {journal} {\bibinfo  {journal} {Proceedings of the National Academy of Sciences}\ }\textbf {\bibinfo {volume} {113}},\ \bibinfo {pages} {3932} (\bibinfo {year} {2016})}\BibitemShut {NoStop}%
\bibitem [{\citenamefont {Schmidt}\ and\ \citenamefont {Lipson}(2009)}]{Schmidt2009}%
  \BibitemOpen
  \bibfield  {author} {\bibinfo {author} {\bibfnamefont {M.}~\bibnamefont {Schmidt}}\ and\ \bibinfo {author} {\bibfnamefont {H.}~\bibnamefont {Lipson}},\ }\bibfield  {title} {\bibinfo {title} {{Distilling Free-Form Natural Laws from Experimental Data}},\ }\href {https://doi.org/10.1126/science.1165893} {\bibfield  {journal} {\bibinfo  {journal} {Science}\ }\textbf {\bibinfo {volume} {324}},\ \bibinfo {pages} {81} (\bibinfo {year} {2009})}\BibitemShut {NoStop}%
\bibitem [{\citenamefont {Raissi}\ \emph {et~al.}(2019)\citenamefont {Raissi}, \citenamefont {Perdikaris},\ and\ \citenamefont {Karniadakis}}]{Raissi2019}%
  \BibitemOpen
  \bibfield  {author} {\bibinfo {author} {\bibfnamefont {M.}~\bibnamefont {Raissi}}, \bibinfo {author} {\bibfnamefont {P.}~\bibnamefont {Perdikaris}},\ and\ \bibinfo {author} {\bibfnamefont {G.}~\bibnamefont {Karniadakis}},\ }\bibfield  {title} {\bibinfo {title} {{Physics-informed neural networks: A deep learning framework for solving forward and inverse problems involving nonlinear partial differential equations}},\ }\href {https://doi.org/10.1016/j.jcp.2018.10.045} {\bibfield  {journal} {\bibinfo  {journal} {Journal of Computational Physics}\ }\textbf {\bibinfo {volume} {378}},\ \bibinfo {pages} {686} (\bibinfo {year} {2019})}\BibitemShut {NoStop}%
\bibitem [{\citenamefont {Karniadakis}\ \emph {et~al.}(2021)\citenamefont {Karniadakis}, \citenamefont {Kevrekidis}, \citenamefont {Lu}, \citenamefont {Perdikaris}, \citenamefont {Wang},\ and\ \citenamefont {Yang}}]{Karniadakis2021}%
  \BibitemOpen
  \bibfield  {author} {\bibinfo {author} {\bibfnamefont {G.~E.}\ \bibnamefont {Karniadakis}}, \bibinfo {author} {\bibfnamefont {I.~G.}\ \bibnamefont {Kevrekidis}}, \bibinfo {author} {\bibfnamefont {L.}~\bibnamefont {Lu}}, \bibinfo {author} {\bibfnamefont {P.}~\bibnamefont {Perdikaris}}, \bibinfo {author} {\bibfnamefont {S.}~\bibnamefont {Wang}},\ and\ \bibinfo {author} {\bibfnamefont {L.}~\bibnamefont {Yang}},\ }\bibfield  {title} {\bibinfo {title} {{Physics-informed machine learning}},\ }\href {https://doi.org/10.1038/s42254-021-00314-5} {\bibfield  {journal} {\bibinfo  {journal} {Nature Reviews Physics}\ }\textbf {\bibinfo {volume} {3}},\ \bibinfo {pages} {422} (\bibinfo {year} {2021})}\BibitemShut {NoStop}%
\bibitem [{\citenamefont {Yazdani}\ \emph {et~al.}(2020)\citenamefont {Yazdani}, \citenamefont {Lu}, \citenamefont {Raissi},\ and\ \citenamefont {Karniadakis}}]{Yazdani2020}%
  \BibitemOpen
  \bibfield  {author} {\bibinfo {author} {\bibfnamefont {A.}~\bibnamefont {Yazdani}}, \bibinfo {author} {\bibfnamefont {L.}~\bibnamefont {Lu}}, \bibinfo {author} {\bibfnamefont {M.}~\bibnamefont {Raissi}},\ and\ \bibinfo {author} {\bibfnamefont {G.~E.}\ \bibnamefont {Karniadakis}},\ }\bibfield  {title} {\bibinfo {title} {{Systems biology informed deep learning for inferring parameters and hidden dynamics}},\ }\href {https://doi.org/10.1371/JOURNAL.PCBI.1007575} {\bibfield  {journal} {\bibinfo  {journal} {PLOS Computational Biology}\ }\textbf {\bibinfo {volume} {16}},\ \bibinfo {pages} {e1007575} (\bibinfo {year} {2020})}\BibitemShut {NoStop}%
\bibitem [{\citenamefont {Lagergren}\ \emph {et~al.}(2020)\citenamefont {Lagergren}, \citenamefont {Nardini}, \citenamefont {Baker}, \citenamefont {Simpson},\ and\ \citenamefont {Flores}}]{Lagergren2020}%
  \BibitemOpen
  \bibfield  {author} {\bibinfo {author} {\bibfnamefont {J.~H.}\ \bibnamefont {Lagergren}}, \bibinfo {author} {\bibfnamefont {J.~T.}\ \bibnamefont {Nardini}}, \bibinfo {author} {\bibfnamefont {R.~E.}\ \bibnamefont {Baker}}, \bibinfo {author} {\bibfnamefont {M.~J.}\ \bibnamefont {Simpson}},\ and\ \bibinfo {author} {\bibfnamefont {K.~B.}\ \bibnamefont {Flores}},\ }\bibfield  {title} {\bibinfo {title} {{Biologically-informed neural networks guide mechanistic modeling from sparse experimental data}},\ }\href {https://doi.org/10.1371/journal.pcbi.1008462} {\bibfield  {journal} {\bibinfo  {journal} {PLOS Computational Biology}\ }\textbf {\bibinfo {volume} {16}},\ \bibinfo {pages} {e1008462} (\bibinfo {year} {2020})}\BibitemShut {NoStop}%
\bibitem [{\citenamefont {Cranmer}\ \emph {et~al.}(2019)\citenamefont {Cranmer}, \citenamefont {Xu}, \citenamefont {Battaglia},\ and\ \citenamefont {Ho}}]{Cranmer2019}%
  \BibitemOpen
  \bibfield  {author} {\bibinfo {author} {\bibfnamefont {M.~D.}\ \bibnamefont {Cranmer}}, \bibinfo {author} {\bibfnamefont {R.}~\bibnamefont {Xu}}, \bibinfo {author} {\bibfnamefont {P.}~\bibnamefont {Battaglia}},\ and\ \bibinfo {author} {\bibfnamefont {S.}~\bibnamefont {Ho}},\ }\bibfield  {title} {\bibinfo {title} {{Learning Symbolic Physics with Graph Networks}},\ }\href {http://arxiv.org/abs/1909.05862} {\bibfield  {journal} {\bibinfo  {journal} {ArXiv}\ } (\bibinfo {year} {2019})},\ \Eprint {https://arxiv.org/abs/1909.05862} {arXiv:1909.05862} \BibitemShut {NoStop}%
\bibitem [{\citenamefont {Rackauckas}\ \emph {et~al.}(2020)\citenamefont {Rackauckas}, \citenamefont {Ma}, \citenamefont {Martensen}, \citenamefont {Warner}, \citenamefont {Zubov}, \citenamefont {Supekar}, \citenamefont {Skinner}, \citenamefont {Ramadhan},\ and\ \citenamefont {Edelman}}]{Rackauckas2020}%
  \BibitemOpen
  \bibfield  {author} {\bibinfo {author} {\bibfnamefont {C.}~\bibnamefont {Rackauckas}}, \bibinfo {author} {\bibfnamefont {Y.}~\bibnamefont {Ma}}, \bibinfo {author} {\bibfnamefont {J.}~\bibnamefont {Martensen}}, \bibinfo {author} {\bibfnamefont {C.}~\bibnamefont {Warner}}, \bibinfo {author} {\bibfnamefont {K.}~\bibnamefont {Zubov}}, \bibinfo {author} {\bibfnamefont {R.}~\bibnamefont {Supekar}}, \bibinfo {author} {\bibfnamefont {D.}~\bibnamefont {Skinner}}, \bibinfo {author} {\bibfnamefont {A.}~\bibnamefont {Ramadhan}},\ and\ \bibinfo {author} {\bibfnamefont {A.}~\bibnamefont {Edelman}},\ }\bibfield  {title} {\bibinfo {title} {{Universal Differential Equations for Scientific Machine Learning}},\ }\href {http://arxiv.org/abs/2001.04385} {\bibfield  {journal} {\bibinfo  {journal} {ArXiv}\ } (\bibinfo {year} {2020})},\ \Eprint {https://arxiv.org/abs/2001.04385} {arXiv:2001.04385} \BibitemShut {NoStop}%
\bibitem [{\citenamefont {Philipps}\ \emph {et~al.}(2024)\citenamefont {Philipps}, \citenamefont {Schmid},\ and\ \citenamefont {Hasenauer}}]{Philipps2024}%
  \BibitemOpen
  \bibfield  {author} {\bibinfo {author} {\bibfnamefont {M.}~\bibnamefont {Philipps}}, \bibinfo {author} {\bibfnamefont {N.}~\bibnamefont {Schmid}},\ and\ \bibinfo {author} {\bibfnamefont {J.}~\bibnamefont {Hasenauer}},\ }\bibfield  {title} {\bibinfo {title} {{Universal differential equations for systems biology: Current state and open problems}},\ }\href {https://doi.org/10.1101/2024.11.29.626122} {\bibfield  {journal} {\bibinfo  {journal} {bioRxiv}\ ,\ \bibinfo {pages} {2024.11.29.626122}} (\bibinfo {year} {2024})}\BibitemShut {NoStop}%
\bibitem [{\citenamefont {Nov{\'{a}}k}\ and\ \citenamefont {Tyson}(2008)}]{Novak2008}%
  \BibitemOpen
  \bibfield  {author} {\bibinfo {author} {\bibfnamefont {B.}~\bibnamefont {Nov{\'{a}}k}}\ and\ \bibinfo {author} {\bibfnamefont {J.~J.}\ \bibnamefont {Tyson}},\ }\bibfield  {title} {\bibinfo {title} {{Design principles of biochemical oscillators}},\ }\href {https://doi.org/10.1038/nrm2530} {\bibfield  {journal} {\bibinfo  {journal} {Nature Reviews Molecular Cell Biology}\ }\textbf {\bibinfo {volume} {9}},\ \bibinfo {pages} {981} (\bibinfo {year} {2008})}\BibitemShut {NoStop}%
\bibitem [{\citenamefont {Prokop}\ \emph {et~al.}(2024)\citenamefont {Prokop}, \citenamefont {Frolov},\ and\ \citenamefont {Gelens}}]{Prokop2024b}%
  \BibitemOpen
  \bibfield  {author} {\bibinfo {author} {\bibfnamefont {B.}~\bibnamefont {Prokop}}, \bibinfo {author} {\bibfnamefont {N.}~\bibnamefont {Frolov}},\ and\ \bibinfo {author} {\bibfnamefont {L.}~\bibnamefont {Gelens}},\ }\bibfield  {title} {\bibinfo {title} {{Enhancing model identification with SINDy via nullcline reconstruction}},\ }\href {https://doi.org/10.1063/5.0199311} {\bibfield  {journal} {\bibinfo  {journal} {Chaos: An Interdisciplinary Journal of Nonlinear Science}\ }\textbf {\bibinfo {volume} {34}},\ \bibinfo {pages} {63135} (\bibinfo {year} {2024})}\BibitemShut {NoStop}%
\bibitem [{\citenamefont {Cebri{\'{a}}n-Lacasa}\ \emph {et~al.}(2024)\citenamefont {Cebri{\'{a}}n-Lacasa}, \citenamefont {Parra-Rivas}, \citenamefont {Ruiz-Reyn{\'{e}}s},\ and\ \citenamefont {Gelens}}]{Cebrian-Lacasa2024}%
  \BibitemOpen
  \bibfield  {author} {\bibinfo {author} {\bibfnamefont {D.}~\bibnamefont {Cebri{\'{a}}n-Lacasa}}, \bibinfo {author} {\bibfnamefont {P.}~\bibnamefont {Parra-Rivas}}, \bibinfo {author} {\bibfnamefont {D.}~\bibnamefont {Ruiz-Reyn{\'{e}}s}},\ and\ \bibinfo {author} {\bibfnamefont {L.}~\bibnamefont {Gelens}},\ }\bibfield  {title} {\bibinfo {title} {{Six decades of the FitzHugh–Nagumo model: A guide through its spatio-temporal dynamics and influence across disciplines}},\ }\href {https://doi.org/10.1016/j.physrep.2024.09.014} {\bibfield  {journal} {\bibinfo  {journal} {Physics Reports}\ }\textbf {\bibinfo {volume} {1096}},\ \bibinfo {pages} {1} (\bibinfo {year} {2024})},\ \Eprint {https://arxiv.org/abs/2404.11403} {arXiv:2404.11403} \BibitemShut {NoStop}%
\bibitem [{Note1()}]{Note1}%
  \BibitemOpen
  \bibinfo {note} {For the FHN model and bicubic model, we normalize such that $u_N\in [-1,1]$ and $v_N\in [-1,1]$ to preserve the system's point symmetry. For other models where all values are non-negative, we normalize to $u_N\in [0,1]$ and $v_N\in [0,1]$. In general, normalization does not affect CLINE’s predictions but improves convergence.}\BibitemShut {Stop}%
\bibitem [{\citenamefont {Messenger}\ \emph {et~al.}(2024)\citenamefont {Messenger}, \citenamefont {Dwyer},\ and\ \citenamefont {Dukic}}]{Messenger2024}%
  \BibitemOpen
  \bibfield  {author} {\bibinfo {author} {\bibfnamefont {D.}~\bibnamefont {Messenger}}, \bibinfo {author} {\bibfnamefont {G.}~\bibnamefont {Dwyer}},\ and\ \bibinfo {author} {\bibfnamefont {V.}~\bibnamefont {Dukic}},\ }\bibfield  {title} {\bibinfo {title} {{Weak-form inference for hybrid dynamical systems in ecology}},\ }\bibfield  {journal} {\bibinfo  {journal} {Journal of the Royal Society Interface}\ }\textbf {\bibinfo {volume} {21}},\ \href {https://doi.org/10.1098/RSIF.2024.0376} {10.1098/RSIF.2024.0376} (\bibinfo {year} {2024})\BibitemShut {NoStop}%
\bibitem [{\citenamefont {Champion}\ \emph {et~al.}(2019)\citenamefont {Champion}, \citenamefont {Brunton},\ and\ \citenamefont {Kutz}}]{Champion2019}%
  \BibitemOpen
  \bibfield  {author} {\bibinfo {author} {\bibfnamefont {K.~P.}\ \bibnamefont {Champion}}, \bibinfo {author} {\bibfnamefont {S.~L.}\ \bibnamefont {Brunton}},\ and\ \bibinfo {author} {\bibfnamefont {J.~N.}\ \bibnamefont {Kutz}},\ }\bibfield  {title} {\bibinfo {title} {{Discovery of Nonlinear Multiscale Systems: Sampling Strategies and Embeddings}},\ }\href {https://doi.org/10.1137/18M1188227} {\bibfield  {journal} {\bibinfo  {journal} {SIAM Journal on Applied Dynamical Systems}\ }\textbf {\bibinfo {volume} {18}},\ \bibinfo {pages} {312} (\bibinfo {year} {2019})}\BibitemShut {NoStop}%
\bibitem [{\citenamefont {{De Boeck}}\ \emph {et~al.}(2021)\citenamefont {{De Boeck}}, \citenamefont {Rombouts},\ and\ \citenamefont {Gelens}}]{DeBoeck2021}%
  \BibitemOpen
  \bibfield  {author} {\bibinfo {author} {\bibfnamefont {J.}~\bibnamefont {{De Boeck}}}, \bibinfo {author} {\bibfnamefont {J.}~\bibnamefont {Rombouts}},\ and\ \bibinfo {author} {\bibfnamefont {L.}~\bibnamefont {Gelens}},\ }\bibfield  {title} {\bibinfo {title} {{A modular approach for modeling the cell cycle based on functional response curves}},\ }\href {https://doi.org/10.1371/journal.pcbi.1009008} {\bibfield  {journal} {\bibinfo  {journal} {PLOS Computational Biology}\ }\textbf {\bibinfo {volume} {17}},\ \bibinfo {pages} {e1009008} (\bibinfo {year} {2021})}\BibitemShut {NoStop}%
\bibitem [{\citenamefont {Parra-Rivas}\ \emph {et~al.}(2023)\citenamefont {Parra-Rivas}, \citenamefont {Ruiz-Reyn{\'{e}}s},\ and\ \citenamefont {Gelens}}]{Parra-Rivas2023}%
  \BibitemOpen
  \bibfield  {author} {\bibinfo {author} {\bibfnamefont {P.}~\bibnamefont {Parra-Rivas}}, \bibinfo {author} {\bibfnamefont {D.}~\bibnamefont {Ruiz-Reyn{\'{e}}s}},\ and\ \bibinfo {author} {\bibfnamefont {L.}~\bibnamefont {Gelens}},\ }\bibfield  {title} {\bibinfo {title} {{Cell cycle oscillations driven by two interlinked bistable switches}},\ }\bibfield  {journal} {\bibinfo  {journal} {Molecular Biology of the Cell}\ }\textbf {\bibinfo {volume} {34}},\ \href {https://doi.org/10.1091/mbc.E22-11-0527} {10.1091/mbc.E22-11-0527} (\bibinfo {year} {2023})\BibitemShut {NoStop}%
\bibitem [{\citenamefont {Strogatz}(2018)}]{strogatz2018nonlinear}%
  \BibitemOpen
  \bibfield  {author} {\bibinfo {author} {\bibfnamefont {S.}~\bibnamefont {Strogatz}},\ }\href@noop {} {\bibinfo {title} {Nonlinear dynamics and chaos: with applications to physics, biology, chemistry, and engineering}} (\bibinfo {year} {2018})\BibitemShut {NoStop}%
\bibitem [{\citenamefont {Gopalsamy}(1992)}]{Gopalsamy1992}%
  \BibitemOpen
  \bibfield  {author} {\bibinfo {author} {\bibfnamefont {K.}~\bibnamefont {Gopalsamy}},\ }\bibfield  {title} {\bibinfo {title} {{Stability and Oscillations in Delay Differential Equations of Population Dynamics}},\ }\bibfield  {journal} {\bibinfo  {journal} {Stability and Oscillations in Delay Differential Equations of Population Dynamics}\ }\href {https://doi.org/10.1007/978-94-015-7920-9} {10.1007/978-94-015-7920-9} (\bibinfo {year} {1992})\BibitemShut {NoStop}%
\bibitem [{\citenamefont {Rombouts}\ \emph {et~al.}(2018)\citenamefont {Rombouts}, \citenamefont {Vandervelde},\ and\ \citenamefont {Gelens}}]{Rombouts2018}%
  \BibitemOpen
  \bibfield  {author} {\bibinfo {author} {\bibfnamefont {J.}~\bibnamefont {Rombouts}}, \bibinfo {author} {\bibfnamefont {A.}~\bibnamefont {Vandervelde}},\ and\ \bibinfo {author} {\bibfnamefont {L.}~\bibnamefont {Gelens}},\ }\bibfield  {title} {\bibinfo {title} {{Delay models for the early embryonic cell cycle oscillator}},\ }\href {https://doi.org/10.1371/JOURNAL.PONE.0194769} {\bibfield  {journal} {\bibinfo  {journal} {PLOS ONE}\ }\textbf {\bibinfo {volume} {13}},\ \bibinfo {pages} {e0194769} (\bibinfo {year} {2018})}\BibitemShut {NoStop}%
\bibitem [{\citenamefont {Glass}\ \emph {et~al.}(2021)\citenamefont {Glass}, \citenamefont {Jin},\ and\ \citenamefont {Riedel-Kruse}}]{Glass2021}%
  \BibitemOpen
  \bibfield  {author} {\bibinfo {author} {\bibfnamefont {D.~S.}\ \bibnamefont {Glass}}, \bibinfo {author} {\bibfnamefont {X.}~\bibnamefont {Jin}},\ and\ \bibinfo {author} {\bibfnamefont {I.~H.}\ \bibnamefont {Riedel-Kruse}},\ }\bibfield  {title} {\bibinfo {title} {{Nonlinear delay differential equations and their application to modeling biological network motifs}},\ }\href {https://doi.org/10.1038/s41467-021-21700-8} {\bibfield  {journal} {\bibinfo  {journal} {Nature Communications}\ }\textbf {\bibinfo {volume} {12}},\ \bibinfo {pages} {1788} (\bibinfo {year} {2021})}\BibitemShut {NoStop}%
\bibitem [{\citenamefont {Rombouts}\ \emph {et~al.}(2023)\citenamefont {Rombouts}, \citenamefont {Verplaetse},\ and\ \citenamefont {Gelens}}]{rombouts2023ups}%
  \BibitemOpen
  \bibfield  {author} {\bibinfo {author} {\bibfnamefont {J.}~\bibnamefont {Rombouts}}, \bibinfo {author} {\bibfnamefont {S.}~\bibnamefont {Verplaetse}},\ and\ \bibinfo {author} {\bibfnamefont {L.}~\bibnamefont {Gelens}},\ }\bibfield  {title} {\bibinfo {title} {The ups and downs of biological oscillators: a comparison of time-delayed negative feedback mechanisms},\ }\href@noop {} {\bibfield  {journal} {\bibinfo  {journal} {Journal of the Royal Society Interface}\ }\textbf {\bibinfo {volume} {20}},\ \bibinfo {pages} {20230123} (\bibinfo {year} {2023})}\BibitemShut {NoStop}%
\bibitem [{\citenamefont {Ikeda}\ \emph {et~al.}(1980)\citenamefont {Ikeda}, \citenamefont {Daido},\ and\ \citenamefont {Akimoto}}]{Ikeda1980}%
  \BibitemOpen
  \bibfield  {author} {\bibinfo {author} {\bibfnamefont {K.}~\bibnamefont {Ikeda}}, \bibinfo {author} {\bibfnamefont {H.}~\bibnamefont {Daido}},\ and\ \bibinfo {author} {\bibfnamefont {O.}~\bibnamefont {Akimoto}},\ }\bibfield  {title} {\bibinfo {title} {{Optical Turbulence: Chaotic Behavior of Transmitted Light from a Ring Cavity}},\ }\href {https://doi.org/10.1103/PhysRevLett.45.709} {\bibfield  {journal} {\bibinfo  {journal} {Physical Review Letters}\ }\textbf {\bibinfo {volume} {45}},\ \bibinfo {pages} {709} (\bibinfo {year} {1980})}\BibitemShut {NoStop}%
\bibitem [{\citenamefont {Boucekkine}\ \emph {et~al.}(2004)\citenamefont {Boucekkine}, \citenamefont {Croix},\ and\ \citenamefont {Licandro}}]{boucekkine2004modelling}%
  \BibitemOpen
  \bibfield  {author} {\bibinfo {author} {\bibfnamefont {R.}~\bibnamefont {Boucekkine}}, \bibinfo {author} {\bibfnamefont {D.~d.~l.}\ \bibnamefont {Croix}},\ and\ \bibinfo {author} {\bibfnamefont {O.}~\bibnamefont {Licandro}},\ }\bibfield  {title} {\bibinfo {title} {Modelling vintage structures with ddes: Principles and applications},\ }\href@noop {} {\bibfield  {journal} {\bibinfo  {journal} {Mathematical Population Studies}\ }\textbf {\bibinfo {volume} {11}},\ \bibinfo {pages} {151} (\bibinfo {year} {2004})}\BibitemShut {NoStop}%
\bibitem [{\citenamefont {Lewis}(2003)}]{Lewis2003}%
  \BibitemOpen
  \bibfield  {author} {\bibinfo {author} {\bibfnamefont {J.}~\bibnamefont {Lewis}},\ }\bibfield  {title} {\bibinfo {title} {{Autoinhibition with transcriptional delay: A simple mechanism for the zebrafish somitogenesis oscillator}},\ }\href {https://doi.org/10.1016/S0960-9822(03)00534-7/ATTACHMENT/4097B886-32CB-4847-9486-7A219A5C8A58/MMC9.PDF} {\bibfield  {journal} {\bibinfo  {journal} {Current Biology}\ }\textbf {\bibinfo {volume} {13}},\ \bibinfo {pages} {1398} (\bibinfo {year} {2003})}\BibitemShut {NoStop}%
\bibitem [{\citenamefont {Tan}\ \emph {et~al.}(2023)\citenamefont {Tan}, \citenamefont {Algar}, \citenamefont {Corr{\^{e}}a}, \citenamefont {Small}, \citenamefont {Stemler},\ and\ \citenamefont {Walker}}]{Tan2023}%
  \BibitemOpen
  \bibfield  {author} {\bibinfo {author} {\bibfnamefont {E.}~\bibnamefont {Tan}}, \bibinfo {author} {\bibfnamefont {S.}~\bibnamefont {Algar}}, \bibinfo {author} {\bibfnamefont {D.}~\bibnamefont {Corr{\^{e}}a}}, \bibinfo {author} {\bibfnamefont {M.}~\bibnamefont {Small}}, \bibinfo {author} {\bibfnamefont {T.}~\bibnamefont {Stemler}},\ and\ \bibinfo {author} {\bibfnamefont {D.}~\bibnamefont {Walker}},\ }\bibfield  {title} {\bibinfo {title} {{Selecting embedding delays: An overview of embedding techniques and a new method using persistent homology}},\ }\href {https://doi.org/10.1063/5.0137223/2881154} {\bibfield  {journal} {\bibinfo  {journal} {Chaos}\ }\textbf {\bibinfo {volume} {33}},\ \bibinfo {pages} {53} (\bibinfo {year} {2023})},\ \Eprint {https://arxiv.org/abs/2302.03447} {arXiv:2302.03447} \BibitemShut {NoStop}%
\bibitem [{\citenamefont {Prokop}\ \emph {et~al.}(2025)\citenamefont {Prokop}, \citenamefont {Billen}, \citenamefont {Frolov},\ and\ \citenamefont {Gelens}}]{gitlab_cline2025}%
  \BibitemOpen
  \bibfield  {author} {\bibinfo {author} {\bibfnamefont {B.}~\bibnamefont {Prokop}}, \bibinfo {author} {\bibfnamefont {J.}~\bibnamefont {Billen}}, \bibinfo {author} {\bibfnamefont {N.}~\bibnamefont {Frolov}},\ and\ \bibinfo {author} {\bibfnamefont {L.}~\bibnamefont {Gelens}},\ }\href@noop {} {\bibinfo {title} {Gitlab repository}},\ \bibinfo {howpublished} {\url{https://gitlab.kuleuven.be/gelenslab/publications/cline-nullcline-recovery.git}} (\bibinfo {year} {2025})\BibitemShut {NoStop}%
\end{thebibliography}%

\end{document}